\documentclass[runningheads]{llncs}
\usepackage{graphicx}
\usepackage{todonotes}
\usepackage{multirow}

\usepackage{array}
\usepackage{tabu}
\newcolumntype{M}[1]{>{\centering\arraybackslash}m{#1}}
\newcolumntype{L}[1]{>{\arraybackslash}m{#1}}

\begin{document}
\title{Overview of BioASQ 2022: The tenth BioASQ challenge on Large-Scale Biomedical Semantic Indexing and Question Answering
% \thanks{Supported by organization x.}
}
    \titlerunning{Overview of BioASQ 2022}

\author{
Anastasios Nentidis\inst{1,2} \and
Georgios Katsimpras\inst{1} \and
Eirini Vandorou\inst{1} \and
Anastasia Krithara\inst{1} \and 
Antonio Miranda-Escalada\inst{3} \and
Luis Gasco\inst{3} \and
Martin Krallinger\inst{3} \and
Georgios Paliouras\inst{1}
}
\authorrunning{A. Nentidis et al.}
\institute{
National Center for Scientific Research ``Demokritos'', Athens, Greece\\
\email{\{tasosnent, gkatsibras, evandorou, akrithara, paliourg\}@iit.demokritos.gr}\\
\and
Aristotle University of Thessaloniki, Thessaloniki, Greece\\ \and
Barcelona Supercomputing Center, Barcelona, Spain\\
\email{\{antonio.miranda, lgasco, martin.krallinger\}@bsc.es}
}
\maketitle              % typeset the header of the contribution
\begin{abstract}
%Advancing the state-of-the-art in large-scale biomedical semantic indexing and question answering is the main focus of the BioASQ challenge. BioASQ organizes respective tasks where different teams develop systems that are evaluated on the same benchmark datasets that represent the real information needs of experts in the biomedical domain. 

This paper presents an overview of the tenth edition of the BioASQ challenge in the context of the Conference and Labs of the Evaluation Forum (CLEF) 2022. BioASQ is an ongoing series of challenges that promotes advances in the domain of large-scale biomedical semantic indexing and question answering. In this edition, the challenge was composed of the three established tasks a, b and Synergy, and a new task named DisTEMIST for automatic semantic annotation and grounding of diseases from clinical content in Spanish, a key concept for semantic indexing and search engines of literature and clinical records. This year, BioASQ received more than 170 distinct systems from 38 teams in total for the four different tasks of the challenge. 
% the ones from PA for task a,b S, plus ~40 from DisTEMIST
As in previous years, the majority of the competing systems outperformed the strong baselines, indicating the continuous advancement of the state-of-the-art in this domain.  

%The evaluation results, similarly to previous years, show a performance gain against the baselines which indicates the continuous improvement of the state-of-the-art in this field.

%In this paper, we present an overview of the ninth edition of the BioASQ challenge, which ran as a lab in the Conference and Labs of the Evaluation Forum (CLEF) 2021.
%BioASQ is a series of challenges aiming at the promotion of systems and methodologies for large-scale biomedical semantic indexing and question answering. 
%To this end, shared tasks are organized yearly since 2012, where different teams develop systems that compete on the same demanding benchmark datasets that represent the real information needs of experts in the biomedical domain. 
%This year, the challenge has been extended with the introduction of a new task on medical semantic indexing in Spanish.
%In total, 34 teams with more than 100 systems participated in the three tasks of the challenge. 
%As in previous years, the results of the evaluation reveal that the top-performing systems managed to outperform the strong baselines, which suggests that state-of-the-art systems keep pushing the frontier of research through continuous improvements. \\

\end{abstract}
\section{Introduction}
Advancing the state-of-the-art in large-scale biomedical semantic indexing and question answering has been the main focus of the BioASQ challenge for more than 10 years. To this end, respective tasks are organized annually, where different teams develop systems that are evaluated on the same benchmark datasets that represent the real information needs of experts in the biomedical domain. Many research teams have participated over the years in these challenges or have profited by its publicly available datasets.

In this paper, we present the shared tasks and the datasets of the tenth BioASQ challenge in 2022, as well as an overview of the participating systems and their performance.
The remainder of this paper is organized as follows. Section~\ref{sec:tasks} presents a general description of the shared tasks, which took place from December 2021 to May 2022, and the corresponding datasets developed for the challenge. 
Followed by this is Section~\ref{sec:participants} that provides a brief overview of the systems developed by the participating teams for the different tasks. 
Detailed descriptions for some of the systems are available in the proceedings of the lab. 
Then, in section~\ref{sec:results}, we focus on evaluating the performance of the systems for each task and sub-task, using state-of-the-art evaluation measures or manual assessment.
% The evaluation of the systems, which was carried out  using state-of-the-art measures or manual assessment, is the last focal point of this paper, with remarks regarding the results of each task. 
The final section concludes the paper by giving some conclusions regarding the 2022 BioASQ challenge.
% The conclusions sum up this year's challenge. 

\section{Overview of the tasks}
\label{sec:tasks}
The tenth edition of the BioASQ challenge consisted of four tasks: (1) a large-scale
biomedical semantic indexing task (task 10a), (2) a biomedical question answering task (task 10b), (3) task on biomedical question answering on the developing problem of COVID-19 (task Synergy), all considering documents in English, and (4) a new medical semantic annotation and concept normalization task in Spanish (DisTEMIST). In this section, we first describe the two established tasks 10a and 10b with focus on differences from previous versions of the challenge~\cite{nentidis2021overview}. For a more detailed description of these tasks the readers can refer to \cite{Tsatsaronis2015}. Additionally, we discuss this year's version of the Synergy task and also present the new DisTEMIST task on medical semantic annotation.

\subsection{Large-scale semantic indexing - task 10a}

\begin{table}[!htb]
        \caption{Statistics on test datasets for task 10a. 
        Due to the early adoption of a new NLM policy for fully automated indexing, the third batch finally consists of a single test set.}\label{tab:a_data}
        \centering
    \begin{tabular}{M{0.1\linewidth}M{0.15\linewidth}M{0.3\linewidth}M{0.3\linewidth}}\hline
        \textbf{Batch} & \textbf{Articles} & \textbf{Annotated Articles} & \textbf{Labels per Article}  \\ \hline
        \multirow{5}{*}{1}        & 9659       & 9450                          & 13.03                           \\
                              & 4531       & 4512                          & 12.00                             \\
                              & 4291       & 4269                          & 13.04                             \\
                              & 4256      & 4192                         & 12.81                            \\
                              & 4862       & 4802                          & 12.75                             \\ \hline

    Total                  & 27599      & 27225                         & 12.72                             \\ \hline
    \multirow{5}{*}{2}        & 8874       & 8818                          & 12.70                             \\
                              & 4071       & 3858                          & 12.38                             \\
                              & 4108       & 4049                          & 12.60                             \\
                              & 3193       & 3045                          & 11.74                             \\
                              & 3078       & 2916                          & 12.07                             \\  \hline

    Total                  & 23324      & 22686                         & 12.29                              \\ \hline
    \multirow{2}{*}{3}        & 2376       & 1870                          & 12.31                             \\
                              & 28       & 0                          & -                             \\ \hline
    Total                  & 2404      & 1870                         & 12.31 \\  \hline
    
    \end{tabular}
\end{table}

In task 10a, participants are asked to classify articles from the PubMed/MEDLINE\footnote{https://pubmed.ncbi.nlm.nih.gov/} digital library into concepts of the MeSH hierarchy. Specifically, new PubMed articles that are not yet annotated by the indexers in the  National Library of Medicine (NLM) are collected to build the test sets for the evaluation of the competing systems. However, NLM scaled-up its policy of fully automated indexing to all MEDLINE citations by mid-2022\footnote{\url{https://www.nlm.nih.gov/pubs/techbull/nd21/nd21\_medline\_2022.html}}. In response to this change, the schedule of task 10a was shifted a few weeks earlier in the year and the task was completed in fewer rounds compared to previous years. The details of each test set are shown in Table \ref{tab:a_data}. In consequence, we believe that, ten years after its initial introduction, task a full-filled its goal in facilitating the advancement of biomedical semantic indexing research and no new editions of this task are planned in the context of the BioASQ challenge.

The task was designed into three independent batches of 5 weekly test sets each. However, due to the early adoption of the new NLM policy the third batch finally consists of a single test set. A second test set was also initially released in the context of the third batch, but due to the fully automated annotation of all its articles by NLM, it was disregarded and no results will be released for it.
Overall, two scenarios are provided in this task: i) on-line and ii) large-scale. The test sets contain new articles from all available journals. 
Similar to previous versions of the task \cite{balikas13}, standard flat and hierarchical information retrieval measures were used to evaluate the competing systems as soon as the annotations from the NLM indexers were available.
Moreover, for each test set, participants had to submit their answers in 21 hours. 
% However, as it has been observed that new MeSH annotations are released in PubMed earlier that in previous years, we shifted the submission period accordingly to avoid having some annotations available from NLM while the task is still running.
Additionally, a training dataset that consists of 16,218,838 articles with 12.68 labels per article, on average, and covering 29,681 distinct MeSH labels in total was provided for task 10a.

\subsection{Biomedical semantic QA - task 10b}

Task 10b consists of a large-scale question answering challenge in which participants have to develop systems for all the stages of question answering in the biomedical domain. As in previous editions, the task examines four types of questions: “yes/no”, “factoid”, “list” and “summary” questions \cite{balikas13}.
%Task 9b aims at providing a realistic large-scale question answering challenge offering to the participating teams the opportunity to develop systems for all the stages of question answering in the biomedical domain. Four types of questions are considered in the task: “yes/no”, “factoid”, “list” and “summary” questions \cite{balikas13}.
In this edition, the available training dataset, which the competing teams had to use to develop their systems, contains 4,234 questions that are annotated with relevant golden elements and answers from previous versions of the task. 
Table \ref{tab:b_data} shows the details of both training and testing sets for task 10b.

\begin{table}[!htb]
        \caption{Statistics on the training and test datasets of task 10b. The numbers for the documents and snippets refer to averages per question.}\label{tab:b_data}
        \centering
        \begin{tabular}{M{0.08\linewidth}M{0.08\linewidth}M{0.09\linewidth}M{0.1\linewidth}M{0.11\linewidth}M{0.14\linewidth}M{0.14\linewidth}M{0.15\linewidth}}\hline
        \textbf{Batch} 	& \textbf{Size} 	&	\textbf{Yes/No}	&\textbf{List}	&\textbf{Factoid}	&\textbf{Summary}& \textbf{Documents} 	& \textbf{Snippets}  	\\ \hline
        Train       &		4,234		&	1148	&816	&1252		&1018	&		9.22			&	12.24			 	\\
        Test 1		&		90			&	23		&14		&34			&19		&		3.22			&	4.06			 	\\
        Test 2		&		90			&	18		&15		&34			&23		&		3.13			&	3.79			 	\\
        Test 3		&		90			&	25		&11		&32			&22		&		2.76			&	3.33			 	\\ 
        Test 4		&		90			&	24		&12		&31			&23		&		2.77			&	3.51			 	\\
        Test 5		&		90			&	28		&18		&29			&15		&		3.01			&	3.60			 	\\
        Test 6		&		37			&	6		&15		&6			&10		&		3.35			&	4.78			 	\\ \hline                    
        \textbf{Total}	&	4,721		&	1272	&901	&1418		&1130	&		3.92			&	5.04			 	\\ \hline 
        
        \end{tabular}
\end{table}

Differently from previous challenges, task 10b was split into six independent bi-weekly batches. These include five official batches, as in previous versions of the task, and an additional sixth batch with questions posed by new biomedical experts. The motivation for this additional batch was to investigate how interesting could be the responses of the systems for biomedical experts that are not familiar with BioASQ. 
In particular, a collaborative schema was adopted for this additional batch, where the new experts posed their questions in the field of biomedicine and the experienced BioASQ expert team reviewed these questions to guarantee their quality. The test set of the sixth batch contains 37 questions developed by eight new experts.

% and the two phases for each batch run during two consecutive days. This year, an additional sixth batch was introduced 
Task 10b is also divided into two phases: (phase A) the retrieval of the required information and (phase B) answering the question, which run during two consecutive days for each batch.
In each phase, the participants receive the corresponding test set and have 24 hours to submit the answers of their systems.
This year, a test set of 90 questions, written in English, was released for phase A and the participants were expected to identify and submit relevant elements from designated resources, including PubMed/MEDLINE articles and snippets extracted from these articles.
Then, the manually selected relevant articles and snippets for these 90 questions were also released in phase B and the participating systems were asked to respond with \textit{exact answers}, that is entity names or short phrases, and \textit{ideal answers}, that is, natural language summaries of the requested information.

\subsection{Task Synergy}
% text taken from BioASQ at CLEF2021: Large-Scale Biomedical Semantic Indexing and Question Answering
In order to make the advancements of biomedical information retrieval and questions answering available for the study of developing problems, we aim at a synergy between the biomedical experts and the automated question answering systems. So that the experts receive and assess the systems' responses and their assessment is fed back to the systems in order to help improving them, in a continuous iterative process.
% , as presented in Figure {\ref{fig:synergy}}.
% \begin{figure*}[!htb]%figure1
% \centerline{\includegraphics[width=0.7\textwidth]{figures/BioASQ_synergy.png}}
% \caption{The iterative dialogue between the experts and the systems in the BioASQ Synergy task on question answering for COVID-19. }\label{fig:synergy}
% \end{figure*}
 In this direction, last year we introduced the BioASQ Synergy task \cite{nentidis2021overview} envisioning a continuous dialog between the experts and the systems. In this model, the experts pose open questions and the systems provide relevant material and answers for these questions. Then, the experts assess the submitted material (documents and snippets) and answers, and provide feedback to the systems, so that they can improve their responses. 
 This process proceeds with new feedback and new predictions from the systems in an iterative way. 
 
 This year, task Synergy took place in four rounds, focusing on unanswered questions for the developing problem of the COVID-19 disease.
 In each round the systems responses and expert feedback refer to the same questions, although some new questions or new modified versions of some questions could be added into the test sets. 
 Table \ref{tab:syn_data} shows the details of the datasets used in task Synergy.

\begin{table}[!htb]
	\caption{Statistics on the datasets of task Synergy. ``Answer'' stands for questions marked as having enough relevant material from previous rounds to be answered. ``Feedback'' stands for questions that already have some expert feedback from previous rounds.}\label{tab:syn_data}
        \centering
        \begin{tabular}{c c c c c c c c}\hline
      	 \textbf{Round}  & \textbf{Size}  & \textbf{Yes/No} &\textbf{List} &\textbf{Factoid} &\textbf{Summary}& \textbf{Answer}  & \textbf{Feedback}   \\
		\hline
		 
 1  &  72   & 21  &20  &13 &18  &	 13  & 26\\
 2  &  70   & 20  &19  &13 &18  &  25   & 70  \\
 3  &  70   & 20  &19  &13 &18  &  41   & 70 \\ 
 4  &  64   & 18  &19  &10 &17  &  47  & 64  \\

	 \hline                 
	\end{tabular}
\end{table}

 Contrary to the task B, this task was not structured into phases, but both relevant material and answers were received together. 
 However, for new questions only relevant material (documents and snippets) is required until the expert considers that enough material has been gathered during the previous round and mark the questions as ``ready to answer". 
 When a question receives a satisfactory answer that is not expected to change, the expert can mark the question as ``closed'', indicating that no more material and answers are needed for it. 

In order to reflect the rapid developments in the field, each round of this task utilizes material from the current version of the COVID-19 Open Research Dataset (CORD-19) \cite{wang2020cord}.
This year the time interval between two successive rounds was extended into three weeks, from two weeks in BioASQ9, to keep up with the release of new CORD-19 versions that were less frequent compared to the previous version of the task.
In addition, apart from PubMed documents of the current CORD-19, CORD-19 documents from PubMed Central and ArXiv were also considered as additional resources of knowledge.  
Similar to task b, four types of questions are examined in the Synergy task: yes/no, factoid, list, and summary, and two types of answers, exact and ideal. Moreover, the assessment of the systems' performance is based on the evaluation measures used in task 10b.

\subsection{Medical semantic annotation in Spanish - DisTEMIST}
The DisTEMIST track \cite{amiranda2022overview} tries to overcome the lack of resources for indexing disease information content in languages other than English, moreover harmonizing concept mentions to controlled vocabularies. SNOMED CT was explicitly chosen to normalize disease mentions for DisTEMIST, because it is a comprehensive, multilingual and widely used clinical terminology \cite{benson2012principles}.

Over the last years, scientific production has increased significantly. And, especially with the COVID-19 health crisis, it has become evident that it is necessary to integrate information from multiple data sources, including biomedical literature and clinical records. Therefore, semantic indexing tools need to efficiently work with heterogeneous data sources to achieve that information integration. But they also need to work beyond data in English, in particular considering publications like clinical case reports, as well as electronic medical records, which are generated in the native language of the healthcare professional/system \cite{amano2021tapping}.

In semantic indexing, certain types of concepts or entities are of particular relevance for researches, clinicians as well as patients alike. For instance, more than 20\% of PubMed search queries are related to diseases, disorders, and anomalies \cite{islamaj2009understanding}, representing the second most used search type after authors. Some efforts were made to extract diseases from text using data in English, like the 2010 i2b2 corpus \cite{i2b2-2010} and NCBI-Disease corpus \cite{ncbi-disease}. Few resource are available for non-English content, particularly with the purpose to process diverse data sources.

DisTEMIST is promoted by the Spanish Plan for the Advancement of Language Technology (Plan TL) \footnote{https://plantl.mineco.gob.es} and organized by the Barcelona Supercomputing Center (BSC) in collaboration with BioASQ. Besides, the extraction of disease mentions is of direct relevance for many use cases such as study of safety issues of biomaterials and implants, or occupational health (associating diseases to professions and occupations).

Figure \ref{fig:distemistoverview} provides an overview of the DisTEMIST shared task setting. Using the generated DisTEMIST resources, participants create their automatic systems and generate predictions. These predictions are later evaluated, and systems are ranked according to their performance. It is structured into two independent sub-tasks (participants may choose to participate in the first, the second, or both), each taking into account a critical scenario:

\begin{figure*}[!htb]%figure2
\centerline{\includegraphics[width=1\textwidth]{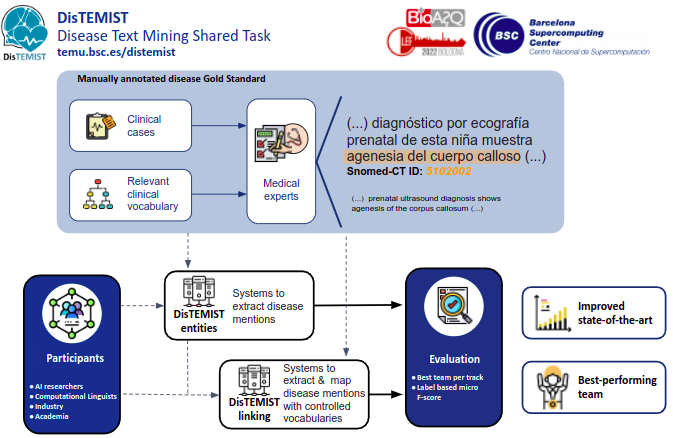}}
\caption{Overview of the DisTEMIST Shared Task.}\label{fig:distemistoverview}
\end{figure*}

\begin{itemize}
    \item \emph{DisTEMIST-entities subtask}. It required automatically finding disease mentions in clinical cases. All disease mentions are defined by their corresponding character offsets (start character and end character) in UTF-8 plain text.
    \item \emph{DisTEMIST-linking subtask}. It is a two-step subtask. It required, first, automatically detection of disease mentions, and then they had to assign, to each mention, its corresponding SNOMED CT concept identifier.
\end{itemize}

To enable the development of disease recognition and linking systems, we have generated the DisTEMIST Gold Standard corpus. It is a collection of 1000 carefully selected clinical cases written in Spanish, that were manually annotated with disease mentions by clinical experts. All mentions were exhaustively revised to mapped them to their corresponding SNOMED CT concept identifier. The manual annotation and code assignment were done following strict annotation guidelines (see the DisTEMIST annotation guidelines in Zenodo\footnote{https://doi.org/10.5281/zenodo.6458078}), and quality checks were implemented. The inter-annotator agreement for the disease mention annotation was 82.3\% (computed as the pairwise agreement between two independent annotators with the 10\% of the corpus).

The corpus was randomly split into training (750 clinical cases) and test (250). Participants used the training set annotations and SNOMED CT assignments for developing their systems, while generating predictions for the test set for evaluation purposes. Table \ref{tab:goldstandardoverview} shows the overview statistics of the DisTEMIST Gold Standard.

\begin{table}
\centering
  \caption{DisTEMIST Gold Standard corpus statistics}
  \label{tab:goldstandardoverview}
  \begin{tabular}{cccccc}
    \toprule
    & Documents & Annotations & Unique codes & Sentences & Tokens\\
    \midrule
    \textbf{Training} & 750 & 8,066 & 4,819 & 12,499 & 305,166 \\
    \textbf{Test} & 250 & 2,599 & 2,484 & 4,179 & 101,152 \\
    \textbf{Total} & 1,000 & 10,665 & 7,303 & 16,678 & 406,318 \\
    \bottomrule
  \end{tabular}
\end{table}

A large number or medical literature is written in languages different from English, this is particularly true for clinical case reports or publications of relevance for health-aspects specific to a certain region or country. For instance, the Scielo repository\footnote{https://scielo.org/} contains 6741148 references in Portuguese and 388528 in Spanish. And relevant non-English-language studies are being published in languages such as Chinese, French, German or Portuguese \cite{amano2021tapping}.

To foster the development of tools also for other languages including low resource languages, we have released DisTEMIST Multilingual Silver Standard corpus. It contains the annotated (and normalized to SNOMED CT) DisTEMIST clinical cases in 6 languages: English, Portuguese, Italian, French, Romanian and Catalan. These resources were generated as follows:

\begin{enumerate}
    \item The original text files are in Spanish. They were translated using a combination of neural machine translation systems to the target languages.
    \item The disease mention annotations were also translated various neural machine translation system.
    \item The translated annotations were transferred to the translated text files. This annotation transfer technology also includes the transfer of the SNOMED CT normalization.
\end{enumerate}

A more in-depth analysis of the DisTEMIST Gold Standard and Multilingual Silver Standard is presented in the DisTEMIST overview paper \cite{amiranda2022overview}. These two resources are freely available at Zenodo\footnote{https://doi.org/10.5281/zenodo.6408476}.

The SNOMED-CT terminology is commonly used in clinical scenarios, but it is less frequent than MeSH or DeCS for literature indexing applications. To help participants used to working with other terminologies, in addition to the manual mappings to SNOMED-CT, we generated cross-mappings to MeSH, ICD-10, HPO, and OMIM through the UMLS Metathesaurus.

Finally, we have generated the DisTEMIST gazetteer, containing official terms and synonyms from the relevant branches of SNOMED CT for the grounding of disease mentions. This was done because SNOMED CT cover different types of information that need to be recorded in clinical records, not just diseases. Indeed, the July 31, 2021 release of the SNOMED CT International Edition included more than 350,000 concepts. In the evaluation phase, mentions whose assigned SNOMED CT term is not included in the versions 1.0, and 2.0 of the DisTEMIST gazetteer were not considered. This resource is accesible at Zenodo\footnote{https://doi.org/10.5281/zenodo.6458114}.

\section{Overview of participation}
\label{sec:participants}
\subsection{Task 10a}
This year, 8 teams participated with a total of 21 different systems in this task. Below, we provide a brief overview of those systems for which a description was available,
% (or was similar to prior years of the challenge) 
stressing their key characteristics. The participating systems along with their corresponding approaches are listed in Table~\ref{tab:a_sys}. 
% Detailed descriptions for some of the systems are available at the proceedings of the workshop.

\begin{table}[!htb]
        \centering
         \caption{Systems and approaches for task 10a. Systems for which no description was available at the time of writing are omitted. }
        \begin{tabular}{M{0.2\linewidth}M{0.8\linewidth}}\hline
        \textbf{System} & \textbf{Approach} \\ \hline
        xlinear, bertMesh & pecos, tf-idf, linear model, BertMesh, PubMedBERT, multilabel attention head\\\hline
        NLM & SentencePiece, CNN, embeddings, ensembles, PubMedBERT\\\hline
        dmiip\_fdu & BertMesh, PubMedBERT, BioBERT, LTR, SVM\\\hline
        D2V\_scalar  &  Doc2Vec, scalar product projection \\\hline

        \end{tabular}
        \label{tab:a_sys}
\end{table}

The team of Wellcome participated in task 10a with two different systems (``\textit{xlinear}'' and ``\textit{bertMesh}''). In particular, the ``\textit{xlinear}'' model is a linear model that uses tf-idf features and it is heavily optimised for fast training and inference, while the  ``\textit{bertMesh}''\footnote{\url{https://huggingface.co/Wellcome/WellcomeBertMesh}} model is a custom implementation based on the BertMesh that utilizes a multilabel attention head and PubMedBERT. The Institut de Recherche en Informatique de Toulouse (IRIT) team  participated with one system, ``\textit{D2V\_scalar}'', which uses Doc2Vec to map textual information into vectors and then applies a scoring mechanism to filter the results.
This year, the National Library of Medicine (NLM) team competed with five systems that followed the same approaches used by the systems in previous versions of the task \cite{rae2021neural}. Finally, the Fudan University team (``\textit{dmiip\_fdu}'') also relied upon existing systems that already participated in the previous version of the task. Their systems are based on a learning to rank approach, where the component methods include both the deep learning based method BERTMeSH, which extends AttentionXML with BioBERT, and traditional SVM based methods.

As in previous versions of the challenge, two systems, developed by NLM to facilitate the annotation of articles by indexers in MEDLINE/PubMed, were available as baselines for the semantic indexing task. The first system is MTI \cite{morkBioasq2014} as enhanced in \cite{zavorin2016} and the second is an extension of it based on features suggested by the winners of the first version of the task \cite{tsoumakasBioasq}.

\subsection{Task 10b}

In task 10b, 20 teams competed this year with a total of 70 different systems for both phases A and B. In particular, 10 teams with 35 systems participated in phase A, while in phase B, the number of participants and systems were 16 and 49 respectively. Six teams engaged in both phases.
An overview of the technologies employed by the teams is provided in Table \ref{tab:b_sys} for the systems for which a description was available. Detailed descriptions for some of the systems are available at the proceedings of the workshop.

\begin{table}[!htb]
        \centering
         \caption{Systems and approaches for task 10b. Systems for which no information was available at the time of writing are omitted.}
        \begin{tabular}{M{0.23\linewidth}M{0.07\linewidth}M{0.7\linewidth}}\hline
        \textbf{Systems} & \textbf{Phase}& \textbf{Approach} \\ \hline
%       %%% PHASE A and B
        %pa & A, B & BM25, BERT, % document retrieval 
        %    Word2Vec, % snippets
        %    SQuAD, PubMedQA, % exact
        %    BioMed-RoBERTa %ideal
        %\\\hline %ITMO team
        bio-answerfinder & A, B &  
            Bio-AnswerFinder, ElasticSearch, Bio-ELECTRA, ELECTRA, %phase A
            BioBERT, SQuAD, wRWMD,  BM25, LSTM, T5%phaseB
        \\\hline % UCSD team
        bioinfo	 & A, B & BM25, ElasticSearch, distant learning, DeepRank, universal weighting passage mechanism (UPWM), PARADE-CNN, PubMedBERT  \\\hline
        
        LaRSA	 & A, B & ElasticSearch, BM25, SQuAD, Marco Passage Ranking, BioBERT, BoolQA, BART  \\\hline
        
        ELECTROBERT	 & A, B & ELECTRA, ALBERT, BioELECTRA, BERT  \\\hline

        %%%% PHASE A ONLY
        %Google & A & BM25, BioBERT, Synthetic  Query  Generation, BERT, reranking  \\\hline
        RYGH & A & BM25, BioBERT, PubMedBERT, T5, BERTMeSH, SciBERT\\\hline
        gsl & A & BM25, BERT, dual-encoder\\\hline
        BioNIR & A & sBERT, distance metrics \\\hline
        KU-systems & B & 
            BioBERT, data augmentation
        \\\hline
        %NCU-IISR & B & BioBERT, logistic regression, LTR \\\hline %NCU

        %%%% PHASE B ONLY
        %UoT & B & BioBERT, multi-task learning, BC2GM \\\hline
        %BioNLPer & B & BioBERT, multi-task learning, NLTK, ScispaCy \\\hline
        %LabZhu & B & BERT, BioBERT, XLNet, SpanBERT, transfer learning, SQuAD, ensembling\\\hline    %Fudan
        MQ & B & tf-idf, sBERT, DistilBERT \\\hline %Word2Vec, BERT, LSTM, Reinforcement Learning (PPO) \\\hline %Diego
        Ir\_sys & B & BERT, SQuAD1.0, SpanBERT, XLNet, PubMedBERT, BioELECTRA, BioALBERT, BART \\\hline
        %CRJ & B & Proximal Policy Optimization (PPO), word2vec, BERT, Reinforcement % MRes, Final BERT, Another ALBERT, Ensemble 
        UDEL-LAB & B & BioM-ALBERT, BioM-ELECTRA, SQuAD \\\hline
        MQU & B & BART, summarization \\\hline
        NCU-IISR/AS-GIS & B & BioBERT, BERTScore, SQuAD, logistic-regression \\\hline
        %sbert & B & Sentence-BERT, BioBERT, SNLI, MutiNLI, multi-task learning, MQU \\\hline
        \end{tabular}
       \label{tab:b_sys}
\end{table}

%The ``\textit{ITMO}'' team participated in both phases of the task experimenting in its ``pa'' systems with  differing solutions across the batches. In general, for document retrieval the systems follow an approach with two stages. First, they identify initial candidate articles based on BM25, and then they re-rank them using variations of BERT~\cite{Devlin2018}, fine-tuned for the binary classification task with the BioASQ dataset and pseudo-negative documents. They extract snippets from the top documents and rerank them using biomedical Word2Vec based on cosine similarity with the question. To extract exact answers they use BERT fine-tuned on SQUAD~\cite{rajpurkar2016squad} and BioASQ datasets and employ a post-processing to split the answer for list questions and additional fine-tuning on PubMedQA~\cite{jin2019pubmedqa} for yes/no questions. 
%Finally, for ideal answers they generate some candidates from the snippets and their sentences and rerank them using the model used for phase A. In the last batch, they also experiment with generative summarization, developing a model based on BioMed-RoBERTa~\cite{gururangan2020don} to improve the readability and consistency of the produced ideal answers. 

The ``\textit{UCSD}'' team competed in both phases of the task with four systems (``\textit{bio-answerfinder}''). For both phases their systems were built upon previously developed systems~\cite{ozyurt2021end}. In phase A, apart from improving tokenization and morphological query expansion facilities, they introduced a relaxation of the greedy ranked keyword based iterative document retrieval for cases where there were no or very few documents, and combined it with a BM25 based retrieval approach on selected keywords. The keywords are ranked with a cascade of LSTM layers. 
%Combining results of the relaxed greedy ranked keyword based iterative document retrieval with BM25 based retrieval on classifier selected keywords 
For phase B, their systems used a T5 based abstractive summarization system instead of the default extractive summarization subsystem.

Another team participating in both phases is the team from the University of Aveiro. Their systems (``\textit{bioinfo}'') relied on their previous transformer-UPWM model \cite{almeida2021universal} and they also experimented with the PARADE-CNN model \cite{li2020parade}. In both systems, they used a fixed PubMedBERT transformer model. Regarding phase B, they tried to answer the yes or no questions by using a simple classifier over a fixed PubMedBERT transformer model.

The team from Mohamed I Uni participated in both phases with the system ``\textit{LaRSA}''. In phase A, they used ElasticSearch with BM25 as a retriever, Roberta-base-fine tuned on SQuAD as a reader, along with a cross-encoder based re-ranker trained on MS Marco Passage Ranking task. In phase B, they used a BioBERT model fine-tuned on SQuAD for both factoid and list questions, while they used a BioBERT fine-tuned on BoolQA and PubMed QA datasets for yes/no questions. For ideal answers they used a BART model fine-tuned on the CNN dataset and the ebmsum corpus.

The BSRC Alexander Fleming team also participated in both phases with four systems in total. Their systems (``\textit{ELECTROBERT}'') are based on a transformer model that combines the replaced token prediction of the ELECTRA system \cite{clark2020electra} with the sentence order prediction used in the ALBERT system \cite{lan2019albert}, and are pre-trained on the 2022 baseline set of all PubMed abstracts provided by the National Library of Medicine and fine-tuned using pairs of relevant and non-relevant question-abstract pairs generated using the BioASQ9 dataset \cite{nentidis2021overview}.

%%%% PHASE A ONLY
In phase A, the ``\textit{RYGH}'' team participated with five systems. They adopted a multi-stage information retrieval system that utilized the BM25 along with several pre-trained models including BioBERT, PubMedBERT and SciBERT.
The Google team competed also in phase A with five systems (``\textit{gsl}''). Their systems are based on a zero-shot hybrid model consisting of two stages: retrieval and re-ranking. The retrieval model is a hybrid of BM25 and a dual-encoder model while the re-ranking is a cross-attention model with ranking loss and is trained using the output of the retrieval model. The TU Kaiserslautern team participated with five systems (``\textit{BioNIR}'') in phase A. Their systems are based on a sBERT sentence transformer which encodes the query and each abstract, sentence by sentence, and it is trained using the BioASQ 10 dataset. They also apply different distance metrics to score and rank the sentences accordingly.

%%%% PHASE B ONLY
% Exact and ideal
In phase B, the ``\textit{KU-systems}'' team participated with five systems.
Their systems are based on a BioBERT backbone architecture that involves also a data augmentation method which relies on a question generation technique.
There were two teams from the  Macquarie University. The first team participated with two systems (``\textit{MQ}'')  in phase B and focused on finding the ideal answers. Their systems used DistilBERT and were trained on the BioASQ10 dataset. The second team competed with two systems (``\textit{MQU}'') which utilized a BART-based abstractive summarization system.

The Fudan University team participated with four systems (``\textit{Ir\_sys}'') in all four types of question answering tasks in phase B. For Yes/no questions, the employed BERT as their backbone and initialized its weights with BioBERT. For Factoid/List questions, they also used a BERT-based model fine-tuned with SQuAD1.0 and BioASQ 10b Factoid/List training datasets. For Summary questions, they adopted BART as the backbone of their model.

The University of Delaware team participated with five systems (``\textit{UDEL-LAB}'') which are based on BioM-Transformers models \cite{alrowili-shanker-2021-biom}. In particular, they used both BioM-ALBERT and BioM-ELECTRA, and this year they investigated three main areas: optimizing the hyper-parameters settings, merging both List and Factoid questions to address the limited size of the Factoid training dataset, and finally, investigating the randomness with Transformers-based models by submitting two identical models with the same hyper-parameters.

The National Central Uni team competed with four systems ``\textit{NCU-IISR/AS-GIS}'' in phase B. For exact answers, they used a pre-trained BioBERT model and took the possible answer list combined with the snippets score generated by Linear Regression model \cite{zhang2021ncu}. For yes/no questions, they used a BioBERT-MNLI model. For factoid and list type, the used a BioBERT - SQuAD model. For ideal answers, they relied on their previous BERT-based model \cite{zhang2021ncu}. To improve their results they replaced ROUGE-SU4 with BERTScore.

As in previous editions of the challenge, a baseline was provided for phase B exact answers, based on the open source OAQA system\cite{yang2016learning}. This system that relies on more traditional NLP and Machine Learning approaches, used to achieve top performance in older editions of the challenge and now serves a baseline. The system is developed based on the UIMA framework. In particular, question and snippet parsing is done with ClearNLP. Then, MetaMap, TmTool \cite{Wei2016}, C-Value and LingPipe \cite{baldwin2003lingpipe} are employed for identifying concept that are retrieved from the UMLS Terminology Services (UTS). Finally, the relevance of concepts, documents and snippets is identified based on some classifier components and some scoring and ranking techniques are also employed.

\subsection{Task Synergy}

In this edition of the task Synergy 6 teams participated submitting the results from 22 distinct systems. 
An overview of systems and approaches employed in this task is provided in Table \ref{tab:syn}, for the systems for which a description was available. More detailed descriptions for some of the systems are available at the proceedings of the workshop.

\begin{table}[!htb]
        \centering
         \caption{Systems and their approaches for task Synergy. Systems for which no description was available at the time of writing are omitted. }
        \begin{tabular}{M{0.2\linewidth}M{0.8\linewidth}}\hline
        \textbf{System} & \textbf{Approach} \\ \hline
        RYGH  & BM25, BioBERT, PubMedBERT, T5, BERTMeSH, SciBERT\\\hline %, RYGH-2, RYGH-3, RYGH-4, RYGH-5
        PSBST & BERT, SQuAD1.0, SpanBERT, XLNet, PubMedBERT, BioELECTRA, BioALBERT, BART \\\hline
        bio-answerfinder & Bio-ELECTRA++, BERT, weighted relaxed word mover's distance (wRWMD), pyserini with MonoT5, SQuAD, GloVe \\\hline %, bio-answerfinder-2
        %AUEB &  BM25, Word2Vec, Graph-Node Embeddings, SciBERT, DL (JPDRMM) \\\hline %-System1 
        MQ & tf-idf, sBERT, DistilBERT  \\\hline %-1, MQ-2, MQ-3, MQ-4, MQ-5
       % system-1, system-2 & sentence vector similarities\\\hline
       bioinfo & BM25, ElasticSearch, distant learning, DeepRank, universal weighting passage mechanism (UPWM), BERT  \\\hline %-0, bioinfo-1, bioinfo-2, bioinfo-3, bioinfo-4
        %NLM & BM25 model, T5, BART    \\\hline %-1, NLM-2, NLM-3, NLM-4, NLM-5
        %pa-synergy & Lucene full-text search, BERT \\\hline
        \end{tabular}
        \label{tab:syn}
\end{table}

The Fudan University (``\textit{RYGH}'', ``\textit{PSBST}'') competed in task Synergy with the same models they used for task 10b. Additionally, they applied a query expansion technique in the preliminary retrieval stage and they used the Feedback data to further fine-tune the model. 

The ``\textit{UCSD}'' team competed in task Synergy with three systems. Their systems (``\textit{bio-answerfinder}'') used the Bio-AnswerFinder end-to-end QA system they had previously developed \cite{ozyurt2021end} with few improvements.

The  Macquarie University team participated with four systems. Their systems (``\textit{MQ}'') retrieved the documents by sending the unmodified question to the search API provided by BioASQ. Then, the snippets were obtained by re-ranking the document sentences based on cosine similarity with the query using two variants: tf-idf, and sBERT. The ideal answers were obtained by sending the top snippets to a re-ranker based on DistilBERT and trained on the BioASQ9b training data.

The University of Aveiro team participated with five systems. Their systems (``\textit{bioinfo}'') are based on their implementation~\cite{almeida2021bioasq} for the previous edition of Synergy, which employs a relevant feedback approach.
Their approach creates a strong baseline using a simple relevance feedback technique, based on tf-idf and the BM25 algorithm.

\subsection{Task DisTEMIST}
The DisTEMIST track received significant interest from a heterogeneous public. 159 teams registered for the task, and 9 of them submitted their predictions from countries such as Mexico, Germany, Spain, Italy, and Argentina. These teams provided 19 systems for DisTEMIST-entities and 15 for DisTEMIST-linking during the task period. Besides, 6 extra systems were submitted post-workshop.

\begin{table}[!htb]
        \centering
         \caption{Systems and approaches for task DisTEMIST. Systems for which no description was available at the time of writing are omitted.}
        \begin{tabular}{M{0.27\linewidth}M{0.05\linewidth}M{0.66\linewidth}}\hline
        \textbf{Team} & \textbf{Ref} & \textbf{Approach} \\ \hline
        PICUSLab & \cite{moscato2022biomedical} & Entities: fine-tuning pre-trained biomedical language model. Linking: pre-trained biomedical language model embeddings similarity \\\hline
        HPI-DHC & \cite{borchert2022HPI} & Entities: based on Spanish Clinical Roberta. Linking: ensemble of a TF-IDF / character-n-gram based approach + multilingual embeddings (SapBERT)\\\hline
        SINAI & \cite{chizhikova2022sinai} & Entities: fine-tuning two different RoBERTA-based models. Linking: biomedical RoBERTa embeddings cosine similarity \\\hline
        Better Innovations Lab \& Norwegian Centre for E-health Research & \cite{marco2022diagnoza} & Entities: fine-tuning Spanish transformer model. Linking: FastText model embeddings Approximate Nearest Neighbour similarity \\\hline
        NLP-CIC-WFU & \cite{tamayo2022mBERT} & Entities: fine-tuned multilingual BERT for token classification and applied simple post-processing to deal with subword tokenization and some punctuation marks\\\hline 
        PU++ & \cite{reyes2022clinical} & Entities: fine-tuning multilingual BERT. Linking: FastText embeddings cosine similarity\\\hline
        Terminología & \cite{jose2022simple} & Use of terminology resources,  NLP preprocess \& lookup\\\hline
        iREL & - & Entities: fine-tuning BiLSTM-CRF with Spanish medical embeddings \\\hline
        Unicage & \cite{neves2022unicage} & Entities: dictionary lookup from several ontologies\\\hline
        \end{tabular}
        \label{tab:distemist_sys}
\end{table}

Table \ref{tab:distemist_sys} describes the general methods used by the participants. Most teams treated DisTEMIST-entities as a NER problem and used pre-trained language models. For DisTEMIST-linking, the most common approach was generating embeddings with the test entities and the ontology terms and applying vector similarity measures, being cosine distance the most popular one. There are interesting variations, such as combining different language models \cite{chizhikova2022sinai} or similarity measures \cite{marco2022diagnoza}, adding post-processing rules \cite{tamayo2022mBERT}, or using dictionary lookup methods \cite{neves2022unicage,jose2022simple}.

For benchmarking purposes, we introduced two baselines systems for the DisTEMIST-entities subtask, (1) DiseaseTagIt-VT: a vocabulary transfer method based on Levenshtein distance, and (2) DiseaseTagIt-Base: a modified BiLSTM-CRF architecture. To create the DisTEMIST-linking baseline, the output from these two systems was fed into a string matching engine to look for similar terminology entries to the entities. DiseaseTagIt-VT obtained a 0.2262 and 0.124 f1-score in DisTEMIST-entities and DisTEMIST-linking, respectively. DiseaseTagIt-Base reached 0.6935 f1-score in DisTEMIST-entities and 0.2642 in DisTEMIST-linking.

\section{Results}
\label{sec:results}

\subsection{Task 10a}

\begin{table*}[!htbp]
\centering
\caption{Average system ranks across the batches of the task 10a. The ranking for Batch 3 is based on the single test set of this batch. A hyphenation symbol (-) indicates insufficient participation in a batch, that is less than four test sets for Batch1 and Batch2. Systems with insufficient participation in all three batches are omitted.  }
%\begin{adjustbox}{totalheight=0.44\textheight}
\begin{tabular}{M{0.3\linewidth}M{0.1\linewidth}M{0.1\linewidth}M{0.1\linewidth}M{0.1\linewidth}M{0.1\linewidth}M{0.1\linewidth}}\hline
\textbf{System} & \multicolumn{2}{c}{\textbf{Batch 1}} & \multicolumn{2}{c}{\textbf{Batch 2}} & \multicolumn{2}{c}{\textbf{Batch 3}} \\ \hline
& MiF & LCA-F & MiF & LCA-F & MiF & LCA-F \\ \cline{2-7}
NLM System 2         & \textbf{1.5}  & \textbf{1.5}  & 4     & 4      & 7  & 7  \\
NLM System 1           & 2    & 2    & 7.5   & 6.625  & 8  & 8  \\
attention\_dmiip\_fdu  & 3.5  & 4    & 2.25  & \textbf{2}      & 2  & 3  \\
deepmesh\_dmiip\_fdu   & 4.25 & 4.75 &\textbf{ 2}     & 2.5    & \textbf{1}  & 2  \\
NLM CNN                & 5.5  & 6.5  & 9.75  & 10.375 & 12 & 12 \\
MTI First Line Index   & 6.75 & 5.5  & 10.5  & 10.25  & 11 & 11 \\
Default MTI            & 7.25 & 6.5  & 10    & 9.75   & 10 & 10 \\
XLinear model          & 8.75 & 8.75 & 13.75 & 13.75  & 16 & 15 \\
Dexstr system          & 10   & 10   & -     & -      & 18 & 17 \\
Plain dict match       & 12.5 & 12.5 & -     & -      & -  & -  \\
deepmesh\_dmiip\_fdu\_ & -    & -    & 3     & 2.5    & 3  & \textbf{1}  \\
dmiip\_fdu             & -    & -    & 4     & 4.25   & 4  & 5  \\
NLM System 4           & -    & -    & 4.75  & 4.75   & 6  & 6  \\
NLM System 3           & -    & -    & 6.25  & 6.625  & 9  & 9  \\
coomat inference       & -    & -    & 7     & 6.5    & 5  & 4  \\
similar to BertMesh    & -    & -    & 12.25 & 12.25  & 13 & 13 \\
BioASQ Filtering       & -    & -    & 13.75 & 14.5   & 15 & 16 \\
ediranknn              & -    & -    & -     & -      & 14 & 14 \\
svm\_baseline          & -    & -    & -     & -      & 17 & 18 \\
\hline
\end{tabular}
%\end{adjustbox}
\label{tab:a_res}
\end{table*}

In task 10a, each of the three batches were independently evaluated as presented in Table~\ref{tab:a_res}.
As in previous editions of the task, the classification performance of the systems was measured with standard evaluation measures \cite{balikas13}, both hierarchical and flat. In particular, the official measures for identifying the winners of each batch were the Lowest Common Ancestor F-measure (LCA-F) and the micro F-measure (MiF) \cite{kosmopoulos2015evaluation}.

As each batch consists of five test sets, we compare the participating systems based on their average rank across all multiple datasets, as suggested by Demšar \cite{Demsar06}. 
Based on the rules of the challenge, the average rank of each system for a batch is the average of the four best ranks of the system in the five test sets of the batch.
However, for the third batch of task 10a, where no multiple test sets are available, the ranking is based in the single available test set. 
In particular, the system with the best performance in a test set gets rank 1.0 for this test set, the second best rank 2.0 and so on. 
In case two or more systems tie, they all receive the average rank.
The average rank of each system, based on both the flat MiF and the hierarchical LCA-F scores, for the three batches of the task are presented in Table~\ref{tab:a_res}. 

\begin{figure*}[!htb]%figure1
\centerline{\includegraphics[width=1\textwidth]{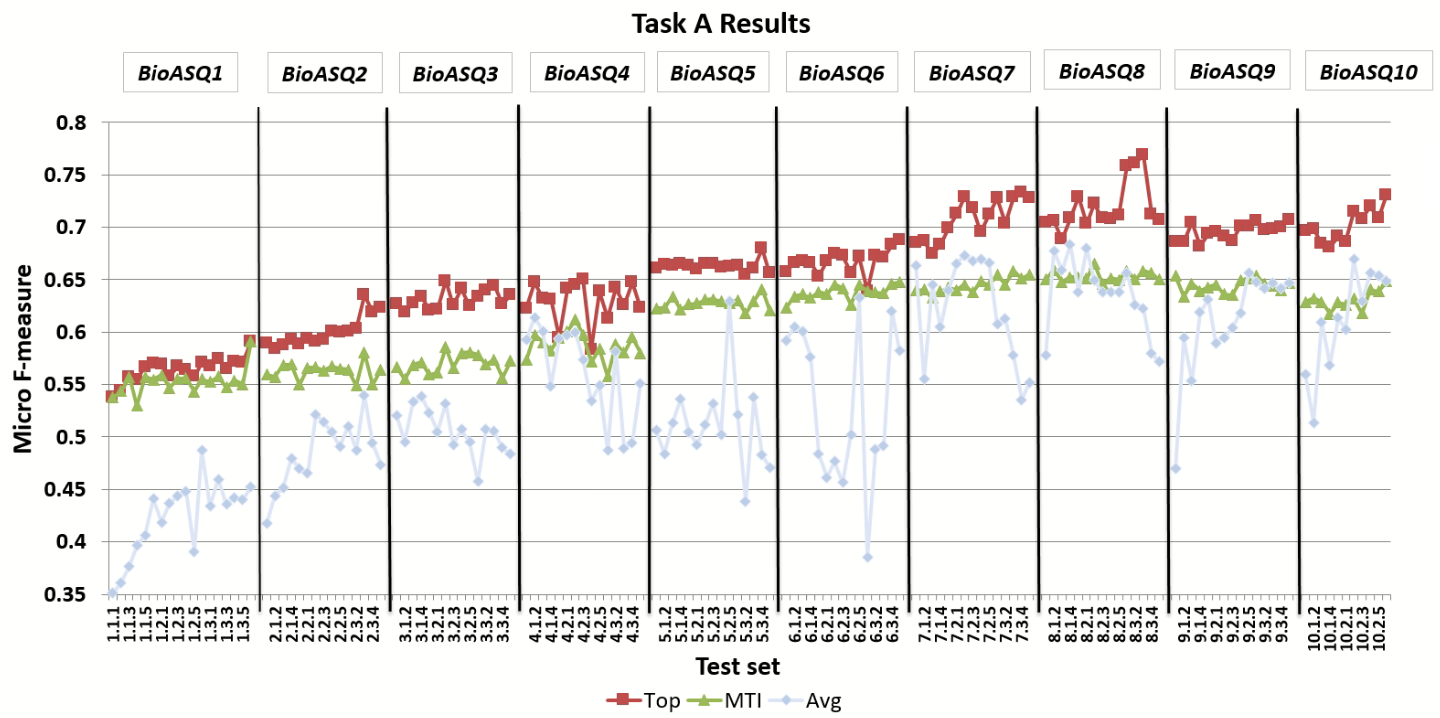}}
\caption{The micro f-measure (MiF) achieved by systems across different years of the BioASQ challenge. For each test set the MiF score is presented for the best performing system (Top) and the MTI, as well as the average micro f-measure of all the participating systems (Avg). }\label{fig:mif}
\end{figure*}

The results of task 10a reveal that several participating systems manage to outperform the strong baselines in all test batches and considering either the flat or the hierarchical measures. 
Namely, the ``NLM'' systems and the ``\textit{dmiip\_fdu}'' systems from the Fudan University team achieve the best performance in all three batches of the task. More detailed results can be found in the online results page\footnote{\footnotesize \url{http://participants-area.bioasq.org/results/10a/}}. 
Figure {\ref{fig:mif}} presents the improvement of the MiF scores achieved by both the MTI baseline and the top performing participant systems through the ten years of the BioASQ challenge.

\subsection{Task 10b}
\textbf{Phase A}: 
In  phase A of task 10b, the evaluation of system performance on document retrieval is based on the Mean Average Precision (MAP) measure. 
For snippets, on the other hand, interpreting the MAP which is based on the number of relevant elements, is more complicated as the same golden snippet may overlap with several distinct submitted snippets. 
Therefore, the official ranking of the systems in snippet retrieval is based on the F-measure, which is calculated based on character overlaps\footnote{\url{http://participants-area.bioasq.org/Tasks/b/eval\_meas\_2021/}}.

Since BioASQ8, a modified version of Average Precision (AP) is adopted for MAP calculation. 
In brief, since BioASQ3, the participant systems are allowed to return up to 10 relevant items (e.g. documents or snippets), and the calculation of AP was modified to reflect this change. However, some questions with fewer than 10 golden relevant items have been observed in the last years, resulting to relatively small AP values even for submissions with all the golden elements. Therefore, the AP calculation was modified to consider both the limit of 10 elements and the actual number of golden elements \cite{nentidis2020overview}.  

\begin{table*}[!htbp]
\centering
\caption{Preliminary results for document retrieval in batch 4 of phase A of task 10b. 
Only the top-10 systems are presented, based on MAP.
}
% \begin{adjustbox}{totalheight=0.4\textheight}
\begin{tabular}{M{0.3\linewidth}M{0.13\linewidth}M{0.13\linewidth}M{0.13\linewidth}M{0.13\linewidth}M{0.12\linewidth}}\hline
\textbf{System} & \textbf{Mean Precision} & \textbf{Mean Recall} & \textbf{Mean F-measure} & \textbf{MAP} & \textbf{GMAP}  \\ \hline
RYGH-3               & \textbf{0.1091}         & 0.5478 & 0.1703    & \textbf{0.4058} & 0.0169 \\
RYGH-1               & 0.1091         & \textbf{0.5496} & \textbf{0.1704}    & 0.4040 & \textbf{0.0183} \\
RYGH                 & 0.1080         & 0.5381 & 0.1684    & 0.3925 & 0.0138 \\
gsl\_zs\_rrf2        & 0.1011         & 0.5024 & 0.1574    & 0.3913 & 0.0083 \\
gsl\_zs\_hybrid      & 0.1011         & 0.5015 & 0.1573    & 0.3904 & 0.0084 \\
RYGH-4               & 0.1111         & 0.5424 & 0.1720    & 0.3883 & 0.0166 \\
RYGH-5               & 0.1100         & 0.5387 & 0.1703    & 0.3873 & 0.0152 \\
gsl\_zs\_rrf1        & 0.0989         & 0.4960 & 0.1541    & 0.3829 & 0.0082 \\
gsl\_zs\_rrf3        & 0.1000         & 0.4997 & 0.1558    & 0.3778 & 0.0089 \\
bioinfo-3            & 0.1133         & 0.5116 & 0.1728    & 0.3613 & 0.0117 \\
% gsl\_zs\_nn          & 0.0933         & 0.4700 & 0.1452    & 0.3609 & 0.0098 \\
% bioinfo-2            & 0.1011         & 0.4708 & 0.1549    & 0.3552 & 0.0083 \\
% bioinfo-4            & 0.1000         & 0.4764 & 0.1539    & 0.3519 & 0.0085 \\
% bioinfo-1            & 0.1011         & 0.4783 & 0.1552    & 0.3502 & 0.0085 \\
% bioinfo-0            & 0.0989         & 0.4825 & 0.1528    & 0.3404 & 0.0077 \\
% LaRSA                & 0.1152         & 0.4631 & 0.1703    & 0.3342 & 0.0073 \\
% bio-answerfinder     & 0.2500         & 0.4180 & 0.2718    & 0.3217 & 0.0064 \\
% bio-answerfinder-3   & 0.2069         & 0.3817 & 0.2317    & 0.3106 & 0.0037 \\
% ELECTROLBERT-2       & 0.1022         & 0.4889 & 0.1574    & 0.3101 & 0.0075 \\
% bio-answerfinder-4   & 0.1383         & 0.4469 & 0.1886    & 0.3056 & 0.0066 \\
% bio-answerfinder-2   & 0.1360         & 0.4335 & 0.1849    & 0.3034 & 0.0060 \\
% AUEB-System2         & 0.0889         & 0.4070 & 0.1362    & 0.3021 & 0.0026 \\
% Basic e2e mid speed  & 0.1503         & 0.4069 & 0.1976    & 0.2924 & 0.0048 \\
% The basic end-to-end & 0.1432         & 0.3854 & 0.1881    & 0.2804 & 0.0034 \\
% AUEB-System1         & 0.0800         & 0.3885 & 0.1238    & 0.2592 & 0.0017 \\
% Deep ML methods for  & 0.0556         & 0.2326 & 0.0834    & 0.1245 & 0.0004 \\
% MindLab QA System    & 0.0556         & 0.2326 & 0.0834    & 0.1245 & 0.0004 \\
\hline
\\
\end{tabular}
% \end{adjustbox}
\label{tab:bA_res_doc}
% \end{table*}

% \begin{table*}[!htbp]
\centering
\caption{Preliminary results for snippet retrieval in batch 4 of phase A of task 10b. Only the top-10 systems are presented, based on F-measure.
        }
\begin{tabular}{M{0.3\linewidth}M{0.14\linewidth}M{0.12\linewidth}M{0.14\linewidth}M{0.12\linewidth}M{0.12\linewidth}}\hline
\textbf{System} & \textbf{Mean Precision} & \textbf{Mean Recall} & \textbf{Mean F-measure} & \textbf{MAP} & \textbf{GMAP}  \\ \hline
bio-answerfinder     & \textbf{0.1270}         & 0.2790 & \textbf{0.1619}    & 0.4905 & 0.0047 \\
RYGH-5               & 0.1126         & 0.3292 & 0.1578    & 0.6596 & 0.0036 \\
RYGH-4               & 0.1119         & \textbf{0.3333 }& 0.1577    & \textbf{0.6606} & 0.0036 \\
bio-answerfinder-3   & 0.1114         & 0.2672 & 0.1463    & 0.4456 & 0.0031 \\
RYGH-3               & 0.0859         & 0.2862 & 0.1257    & 0.3669 & \textbf{0.0067} \\
RYGH-1               & 0.0845         & 0.2801 & 0.1235    & 0.3620 & 0.0059 \\
RYGH                 & 0.0836         & 0.2747 & 0.1215    & 0.3523 & 0.0049 \\
bio-answerfinder-4   & 0.0887         & 0.2342 & 0.1197    & 0.2973 & 0.0031 \\
Basic e2e mid speed  & 0.0887         & 0.2146 & 0.1184    & 0.3321 & 0.0019 \\
bio-answerfinder-2   & 0.0878         & 0.2301 & 0.1182    & 0.2949 & 0.0031 \\
% The basic end-to-end & 0.0850         & 0.2089 & 0.1140    & 0.3225 & 0.0013 \\
% AUEB-System2         & 0.0657         & 0.1648 & 0.0905    & 0.2653 & 0.0007 \\
% AUEB-System1         & 0.0597         & 0.1597 & 0.0834    & 0.2480 & 0.0008 \\
% LaRSA                & 0.0553         & 0.1312 & 0.0720    & 0.1393 & 0.0009 \\
% Deep ML methods for  & 0.0121         & 0.0365 & 0.0163    & 0.0363 & 0.0000 \\
% MindLab QA System    & 0.0121         & 0.0374 & 0.0163    & 0.0358 & 0.0000 \\
        \hline
        \\
        \end{tabular}
        \label{tab:bA_res_sni}
% \end{table*}
% \begin{table*}[!htbp]

\centering
\caption{Results for batch 4 for exact answers in phase B of task 10b.
Only the top-10 systems based on Yes/No F1 and the BioASQ Baseline are presented.}
\begin{tabular}
{M{0.205\linewidth}M{0.0852\linewidth}M{0.0852\linewidth}M{0.105\linewidth}M{0.11\linewidth}M{0.0852\linewidth}M{0.0852\linewidth}M{0.0852\linewidth}M{0.0852\linewidth}}
\hline

\textbf{System} & \multicolumn{2}{c}{\textbf{Yes/No}} & \multicolumn{3}{c}{\textbf{Factoid}} & \multicolumn{2}{c}{\textbf{List}} \\ 
\hline
& F1 & Acc. & Str. Acc. & Len. Acc. & MRR & Prec. & Rec. & F1 \\ \cline{2-9}    
UDEL-LAB3          & \textbf{1.0000}      & \textbf{1.0000}      & 0.5161      & 0.6129       & 0.5484 & 0.5584     & 0.4438 & 0.4501    \\
UDEL-LAB4          & \textbf{1.0000}     & \textbf{1.0000}      & 0.5484      & 0.6129       & 0.5613 & \textbf{0.6162}     & 0.4753 & 0.4752    \\
UDEL-LAB5          &\textbf{1.0000}       & \textbf{1.0000}      & 0.5161      & 0.5806       & 0.5484 & 0.6132     & 0.4426 & 0.4434    \\
Ir\_sys1           &\textbf{1.0000}      & \textbf{1.0000}      & 0.4839      & 0.6452       & 0.5495 & 0.4444     & 0.2410 & 0.2747    \\
Ir\_sys2           & \textbf{1.0000}       & \textbf{1.0000}      & 0.4516      & 0.5161       & 0.4839 & 0.3889     & 0.2847 & 0.2718    \\
lalala             &\textbf{1.0000}      & \textbf{1.0000 }     & \textbf{0.5806}      & 0.6452       & \textbf{0.5995} & 0.4089     & 0.4507 & 0.3835    \\
UDEL-LAB1          & 0.9515       & 0.9583      & 0.4839      & 0.6129       & 0.5387 & 0.5799     & 0.5017 & 0.4950    \\
UDEL-LAB2          & 0.9515       & 0.9583      & 0.4839      & 0.6129       & 0.5484 & 0.5834     & \textbf{0.5844} & \textbf{0.5386}    \\
bio-answerfinder   & 0.9473       & 0.9583      & 0.3548      & 0.4194       & 0.3871 & 0.3727     & 0.2701 & 0.2733    \\
BioASQ\_Baseline   & 0.2804       & 0.2917      & 0.1613      & 0.3226       & 0.2177 & 0.2163     & 0.4035 & 0.2582   \\
\hline
\end{tabular}
%\end{adjustbox}

% ranked by the F1 score for list questions.
\label{tab:bB_res}
\end{table*}

Tables \ref{tab:bA_res_doc} and \ref{tab:bA_res_sni} present some indicative preliminary results for the retrieval of documents and snippets in batch 4. The full results are available online in the result page of task 10b, phase A\footnote{\footnotesize \url{http://participants-area.bioasq.org/results/10b/phaseA/}}. These results are currently preliminary, as the manual assessment of the system responses by the BioASQ team of biomedical experts is still in progress and the final results for the task 10b will be available after its completion. 
% Figure {\ref{fig:IR}}, which depicts the top system performance during the nine years of BioASQ, reveals that despite the improvements observed, there is still a lot of room for improvement in document and snippet retrieval.

\textbf{Phase B}: 
In phase B of task 10b, the participating systems are expected to submit both exact and ideal answers. 
Regarding the sub-task of ideal answer generation, the BioASQ experts assess all the systems responses, assigning manual scores to each ideal answer~\cite{balikas13}. Then, the official system ranking is based on these manual scores. 
For exact answers, the participating systems are ranked based on their average ranking in the three question types where exact answers are required, excluding summary questions for which no exact answers are submitted.
For list questions the ranking is based on mean F1-measure, for factoid questions on mean reciprocal rank (MRR), and for yes/no questions on the F1-measure, macro-averaged over the classes of yes and no. 
Table~\ref{tab:bB_res} presents some indicative preliminary results on exact answer extraction from batch 4. The full results of phase B of task 10b are available online\footnote{\footnotesize \url{http://participants-area.bioasq.org/results/10b/phaseB/}}. These results are preliminary, as the final results for task 10b will be available after the manual assessment of the system responses by the BioASQ team of biomedical experts.

% \begin{figure*}[!htbp]%figure1
% \centerline{\includegraphics[width=1\textwidth]{figures/9bBA.png}}
% \caption{
% The official evaluation scores of the best performing systems in Task B, Phase A, Information Retrieval, across the nine years of the BioASQ challenge.
% % Since BioASQ3 the participants are allowed to submit up to 10 documents or snippets, therefore the MAP implementation is adjusted accordingly.
% % The results for BioASQ9 are preliminary.
% }\label{fig:IR}
% \end{figure*}

\begin{figure*}[!htbp]%figure1
\centerline{\includegraphics[width=1\textwidth]{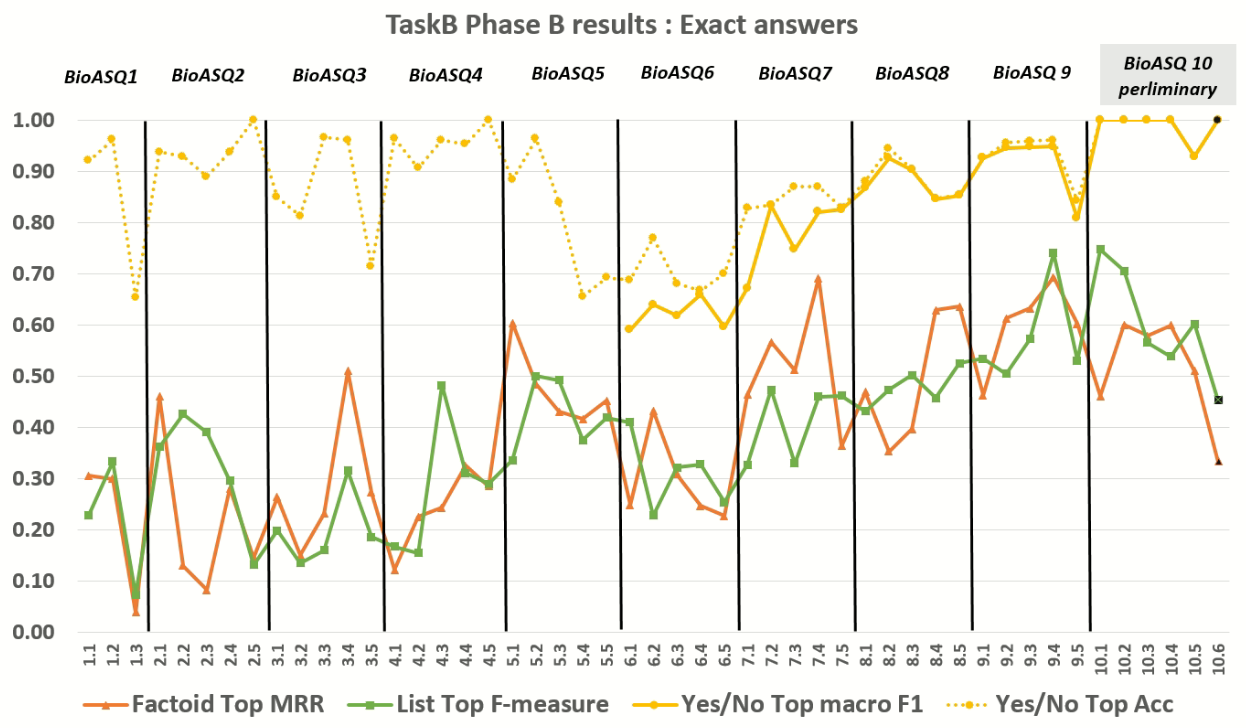}}
\caption{
The evaluation scores of the best performing systems in task B, 
Phase B, for exact answers, across the ten years of the BioASQ
challenge. Since BioASQ6 the official measure for Yes/No questions is the
macro-averaged F1 score (macro F1), but accuracy (Acc) is also presented as the former official measure. The black dots in 10.6 highlight that these scores are for the additional batch with questions from new experts.
% Since BioASQ3 the participants are allowed to submit up to 10 documents or snippets, therefore the MAP implementation is adjusted accordingly.
% The results for BioASQ9 are preliminary.
}\label{fig:Exact}
\end{figure*}

The top performance of the participating systems in exact answer generation for each type of question during the ten years of BioASQ is presented in Figure {\ref{fig:Exact}}.
These preliminary results reveal that the participating systems keep improving in all types of questions.
In batch 4, for instance, presented in Table \ref{tab:bB_res}, in yes/no questions several systems manage to answer correctly for all yes/no questions.
Some improvements are also observed in the preliminary results for list questions compared to the previous years, but there is still more room for improvement, as dose for factoid questions where the preliminary performance is comparable to the one of the previous year.
% In general, Figure {\ref{fig:Exact}} suggests that for the latter types of question there is still more room for improvement.  
The performance of the best systems in the additional collaborative batch (10.6 in Figure {\ref{fig:Exact}}), although clearly decreased compared to the other batches, it is still comparable to them. 
This suggests that the responses of the participating systems could be useful to biomedical experts that are not necessarily familiarized with the BioASQ framework and did not contribute with any questions to the development of the training dataset.

\subsection{Task Synergy}

In task Synergy the participating systems were expected to retrieve documents and snippets, as in phase A of task 10b, and, at the same time, provide answers for some of these questions, as in phase B of task 10b. 
In contrast to task 10b, it is possible that no answer exists for some questions. Therefore only some of the questions provided in each test set, that were indicated to have enough relevant material gathered from previous rounds, require the submission of exact and ideal answers.
Also in contrast to task B, for new questions no golden documents and snippets were provided, while for questions from previous rounds a separate file with feedback from the experts, based on the previously submitted responses, was provided.

The feedback concept was introduced in this task to further assist the collaboration between the systems and the BioASQ team of biomedical experts. The feedback includes the already judged documentation and answers along with their evaluated relevancy to the question. The documents and snippets included in the feedback are not considered valid for submission in the following rounds, and even if accidentally submitted, they were not be taken into account for the evaluation of that round. The evaluation measures for the retrieval of documents and snippets are the MAP and F-measure respectively, as in phase A of task 10b.

Regarding the ideal answers, the systems are ranked according to manual scores assigned to them by the BioASQ experts during the assessment of systems responses as in phase B of task B~\cite{balikas13}. 
For the exact answers, which are required for all questions except the summary ones, the measure considered for ranking the participating systems depends on the question type. 
% Results for Synergy task version 1 are seen in short 
For the yes/no questions, the systems were ranked according to the macro-averaged F1-measure on prediction of no and yes answer. 
For factoid questions, the ranking was based on mean reciprocal rank (MRR) and for list questions on mean F1-measure.

\begin{table}
    \centering
     \caption{Results for document retrieval of the third round of the Synergy 10 task. Only the top-10 systems are presented.}
    \begin{tabular}{M{0.3\linewidth}M{0.14\linewidth}M{0.12\linewidth}M{0.14\linewidth}M{0.12\linewidth}M{0.12\linewidth}}
    \hline
        \textbf{System} & \textbf{Mean precision} & \textbf{Mean Recall} & \textbf{Mean F-Measure} & \textbf{MAP} & \textbf{GMAP} \\ \hline
bio-answerfinder-3 & \textbf{0.3063}         & \textbf{0.2095} & \textbf{0.1970}    & \textbf{0.2622} & \textbf{0.0184} \\
RYGH               & 0.2500         & 0.2017 & 0.1733    & 0.2125 & 0.0157 \\
RYGH-3             & 0.2361         & 0.1699 & 0.1551    & 0.2019 & 0.0079 \\
RYGH-1             & 0.2267         & 0.1677 & 0.1522    & 0.1944 & 0.0116 \\
RYGH-4             & 0.2125         & 0.1642 & 0.1450    & 0.1797 & 0.0105 \\
bio-answerfinder-2 & 0.2484         & 0.1204 & 0.1382    & 0.1736 & 0.0036 \\
bio-answerfinder   & 0.2402         & 0.1187 & 0.1334    & 0.1669 & 0.0031 \\
PSBST2             & 0.1844         & 0.1578 & 0.1327    & 0.1511 & 0.0131 \\
RYGH-5             & 0.1891         & 0.1615 & 0.1353    & 0.1413 & 0.0108 \\
bioinfo-3          & 0.1798         & 0.1060 & 0.1010    & 0.1158 & 0.0025 \\
        \hline 
    \end{tabular}
   \label{tab:synergy1-res}
\end{table}

Some indicative results for the Synergy task are presented in Table~\ref{tab:synergy1-res}.
% and~\ref{tab:synergy1-resSnip} respectively. 
The full results of Synergy task are available online\footnote{\footnotesize \url{http://participants-area.bioasq.org/results/synergy\_v2022/}}. 
% Figure {\ref{fig:synergy-v1-exact}} presents the performance of the best systems in each round of the first version of the the Synergy, revealing that the extraction of exact answers seems to improve during the rounds, particularly for yes/no and factoid questions. 
% As regards the extraction of exact answers, despite the moderate scores in list and factoid questions the experts found useful the submissions of the participants, as most of them (more than 70\%) stated they would be interested in using a tool following the BioASQ Synergy process to identify interesting material and answers for their research.
Although the scores on information retrieval and extraction of exact answers are quite moderate, compared to task 10b, where the questions are not open nether for a developing issue, the experts did found the submissions of the participants useful, as most of them stated they would be interested in using a tool following the BioASQ Synergy process to identify interesting material and answers for their research.

\subsection{Task DisTEMIST}

The performance range of DisTEMIST participants varies depending on the method employed, the subtask (DisTEMIST-entities vs. DisTEMIST-linking), and even within the same team. The highest micro-average F1-score in DisTEMIST-entities is 0.777, and it is 0.5657 in DisTEMIST-linking.

% \usepackage{multirow}
% \usepackage{graphicx}
% \usepackage[normalem]{ulem}
\begin{table}
\centering
\caption{Results of DisTEMIST systems. The best run per team and subtask is shown. For full results, see the DisTEMIST overview paper \cite{amiranda2022overview}. MiP, MiR and MiF stands for micro-averaged Precision, Recall and F1-score. DisTEMIST-e stands for DisTEMIST-entities and DisTEMIST-l stands for DisTEMIST-linking.}
\begin{tabular}{clccccccc}
\multicolumn{1}{l}{} & \multicolumn{1}{l}{} & \multicolumn{3}{c}{\textbf{DisTEMIST-e}} & & \multicolumn{3}{c}{\textbf{DisTEMIST-l}}\\ \hline
\textbf{Team} & \textbf{System} & \textbf{MiP} & \textbf{MiR} & \textbf{MiF} & & \textbf{MiP} & \textbf{MiR} & \textbf{MiF} \\ \hline
\multirow{2}{*}{PICUSLab} & NER\_results & 0.7915 & 0.7629 & 0.777 & & & \\
& EL\_results & & & & & 0.2814 & 0.2748 & 0.278 \\ \hline

\multirow{2}{*}{HPI-DHC} & 3-r.c.e.-linear-lr-pp. & 0.7434 & 0.7483 & 0.7458 & & & \\
& 5-ensemble-reranking-pp. & & & & & 0.6207 & 0.5196 & 0.5657 \\\hline

\multirow{2}{*}{SINAI} & run2-biomedical\_model & 0.752 & 0.7259 & 0.7387 & & & \\
& run1-clinical\_model & & & & & 0.4163 & 0.4081 & 0.4122 \\\hline

\multirow{2}{*}{\shortstack{Better Innovations\\ Lab \& Norwegian.}} & run1-ner & 0.7724 & 0.6925 & 0.7303 & & & \\
& run1-snomed & & & & & 0.5478 & 0.4577 & 0.4987 \\\hline

\multirow{1}{*}{NLP-CIC-WFU} & System\_mBERT & 0.6095 & 0.4938 & 0.5456 & & & \\\hline

\multirow{2}{*}{PU++} & run2\_mbertM5 & 0.601 & 0.4488 & 0.5139 & & & \\
& run2-scieloBERT & & & & & 0.2754 & 0.1494 & 0.1937 \\\hline

\multirow{2}{*}{Terminología} & distemist-subtrack1 & 0.5622 & 0.3772 & 0.4515 & & & \\
& distemist-subtrack2 & & & & & 0.4795 & 0.2292 & 0.3102 \\\hline

\multirow{1}{*}{iREL} & iREL & 0.4984 & 0.3576 & 0.4164 & & & \\\hline

\multirow{1}{*}{Unicage} & XL\_LEX\_3spc & 0.2486 & 0.3303 & 0.2836 & & & \\\hline

\multirow{2}{*}{BSC baselines} & DiseaseTagIt-VT & 0.1568 & 0.4057 & 0.2262 & & 0.1003 & 0.1621 & 0.124 \\
& DiseaseTagIt-Base & 0.7146 & 0.6736 & 0.6935 & & 0.3041 & 0.2336 & 0.2642 \\\hline

\end{tabular}
\label{tab:distemist_results}
\end{table}

As shown in Table \ref{tab:distemist_results}, the top performer in DisTEMIST-entities was the NER system of PICUSLab, based on the fine-tuning of a pre-trained biomedical transformer language model. In the case of DisTEMIST-linking, the highest micro-average F1-score, precision and recall were obtained by the \emph{ensemble-reranking-postprocess} system from the HPI-DHC team. It is based on an ensemble of a TF-IDF and character-n-gram-based approach with multilingual embeddings. Comparing the participant performances with the baseline, all teams outperformed the vocabulary transfer baseline (DiseaseTagIt-VT) in both subtasks. Based on BiLSTM-CRF architecture (DiseaseTagIt-Base), the competitive baseline ranked 5th in DisTEMIST-entities and 6th in DisTEMIST-linking. 
\begin{figure*}
\centerline{\includegraphics[width=1\textwidth]{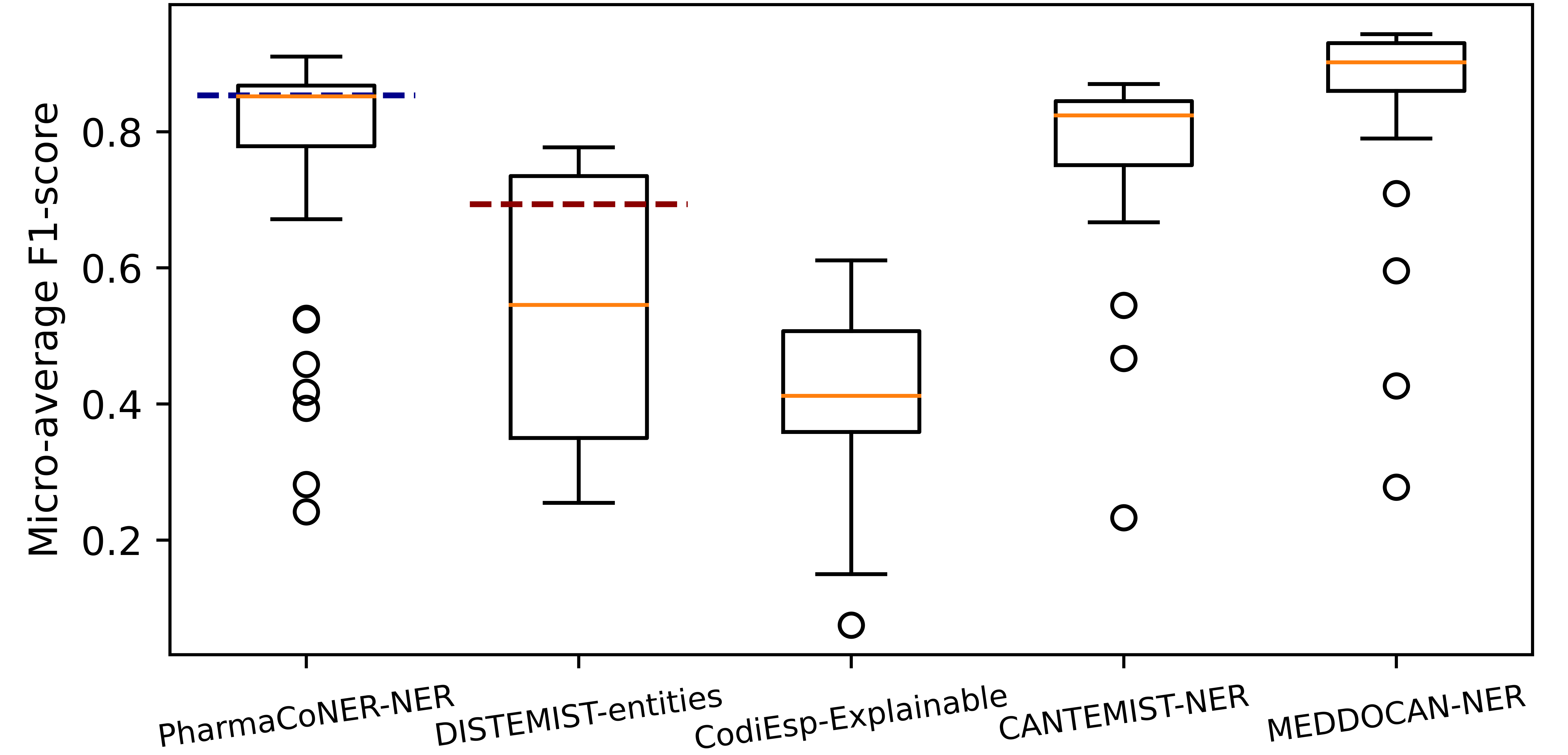}}
\caption{Micro-average F1-score distribution of PharmaCoNER, DisTEMIST, CodiEsp, CANTEMIST, and MEDDOCAN NER systems. Themicro-average F1-scores of PharmaCoNER and DisTEMIST-entities baseline are shown in blue and red, respectively.}\label{fig:pharmaconerdistemist}
\end{figure*}

The DisTEMIST-entities shared task results are comparable to the results of previous shared tasks such as PharmaCoNER \cite{gonzalez2019pharmaconer}, CodiEsp \cite{miranda2020overview}, CANTEMIST \cite{miranda2020named} and MEDDOCAN. All of them are NER challenges on Spanish clinical documents. The results are specially comparable as DisTEMIST-entities, PharmaCoNER's first subtask, and CodiEsp's ExplainableAI subtask used the same corpus of documents, each annotated with different criteria and entity types - DisTEMIST with disease entities, PharmaConer with medication entities, and CodiEsp with Diagnosis and Procedures according to the ICD-10 definitions-. Additionally, the baseline of DisTEMIST and PharmaCoNER used the same architecture.

When comparing the approaches, DisTEMIST NER participants mainly employed combinations of large pre-trained transformer models, and the same is true for highest-scoring CANTEMIST participants. PharmaCoNER took place when these models were unavailable, and the most popular deep learning architectures had Recurrent and Convolutional Neural Networks at their cores. Finally, several successful CodiEsp teams used lexical lookup approaches to match the ICD-10 definitions. PharmaCoNER, CANTEMIST, and MEDDOCAN results were higher -Figure \ref{fig:pharmaconerdistemist} compares the distribution of f1-scores in all tasks-.

This performance difference is directly related to the complexity of the target entities. For instance, the average number of characters per annotation in the PharmaCoNER training set was 9.7 and in DisTEMIST-entities was 24.6. The entity complexity influenced the annotation (the inter-annotator agreement of PharmaCoNER corpus was 93\%, and it was 82.3\% for DisTEMIST) and the system performance. It is remarkable that, given this increase in entity complexity, participants still developed competitive systems in DisTEMIST.

\section{Conclusions}
\label{sec:conclusion}
% An interesting observation in these results is that different systems perform best in different metrics, which reflect different priorities regarding the information retrieval. 

An overview of the tenth BioASQ challenge is provided in this paper. This year, the challenge consisted of four tasks: The two tasks on biomedical semantic indexing and question answering in English, already established through the previous nine years of the challenge, the 2022 version of the Synergy task on question answering for COVID-19, and the new task DisTEMIST on retrieving disease information from medical content in Spanish.
This year, task 10a was completed earlier than expected, due to the early adoption of the new policy of NLM for fully automated indexing. Although a slight trend towards improved scores can be still observed in the results of this tenth year, we believe that the task has successfully completed its main goal, concluding its life cycle.  

The preliminary results for task 10b reveal some improvements in the performance of the top participating systems, mainly for yes/no and list answer generation. However, room for improvement is still available, particularly for factoid and list questions.
The introduction of an additional collaborative batch with questions from new biomedical experts that are not familiarized with BioASQ and did not contribute to the development of the BioASQ datasets before, allows for interesting observations. 
Although the scores for list and factoid answers in this batch are lower compared to the other batches, they are still comparable, highlighting the usefulness of the systems for any biomedical expert. For yes/no questions, in particular, some systems even managed to answer all the questions correctly.

The new task DisTEMIST introduced two new challenging subtasks beyond the one on medical literature. Namely, Named Entity Recognition and Entity Linking of diseases in Spanish clinical documents. Due to the importance of semantic interoperability across data sources, SNOMED CT was the target terminology employed in this task, and multilingual annotated resources have been released. This novel task on disease information indexing in Spanish highlighted the importance of generating resources to develop and evaluate systems that (1) effectively work in multilingual and non-English scenarios and (2) combine heterogeneous data sources.

The second year of the Synergy task in an effort to enable a dialogue between the participating systems with biomedical experts revealed that state-of-the-art systems, despite they still have room for improvement, can be a useful tool for biomedical experts that need specialized information in the context of developing problems such as the COVID-19 pandemic.

As already observed during the last years, the participating systems focus more and more on  deep neural approaches. Almost all competing solutions are based on state-of-the-art neural architectures (BERT, PubMedBERT, BioBERT, BART etc.) adapted to the biomedical domain and specifically to the tasks of BioASQ. New promising approaches have been explored this year, especially for the exact answer generation, leading to improved results.  

Overall, several systems managed to outperform the strong baselines on the challenging tasks offered in BioASQ, as in previous versions of the challenge, and the top preforming of them were able to improve over the state of the art performance from previous years.
BioASQ keeps pushing the research frontier in biomedical semantic indexing and question answering for ten years now, offering both well established and new tasks. 
Lately is has been extended beyond the English language and biomedical literature, with the tasks MESINESP and DisTEMIST. 
In addition, BioASQ reaches a more and more broad community of biomedical experts that may benefit from the advancements in the field. This has been done initially through BioASQ Synergy for COVID-19 and this year with the collaborative batch of task 10b.
The future plans for the challenge include to further extend of the benchmark data for question answering though a community-driven process, as well as extending the Synergy task into other developing problems beyond COVID-19.

\section{Acknowledgments}
Google was a proud sponsor of the BioASQ Challenge in 2021. The tenth edition of BioASQ is also sponsored by the Atypon Systems inc. 
BioASQ is grateful to NLM for providing the baselines for task 10a and to the CMU team for providing the baselines for task 10b.
The DisTEMIST track was supported by the Spanish Plan for advancement of Language Technologies (Plan TL) and the Secretaría de Estado de Digitalización e Inteligencia Artificial (SEDIA), the European Union’s Horizon Europe Coordination \& Support Action under Grant Agreement No 101058779 and by the AI4PROFHEALTH (PID2020-119266RA-I00).
\bibliographystyle{splncs04}
\bibliography{BioASQ8.bib}

\end{document}